\documentclass[10pt,twocolumn]{article}

\usepackage{cvpr}
\usepackage{times}
\usepackage{epsfig}
\usepackage{graphicx}
\usepackage{amsmath}
\usepackage{amssymb}

\usepackage{dsfont}

\usepackage{microtype}
\def\etal{\textit{et al. }}
\def\eg{\textit{e.g., }}

\usepackage{pdfrender}
\newcommand*{\boldcheckmark}{%
  \textpdfrender{
    TextRenderingMode=FillStroke,
    LineWidth=.5pt, 
  }{\checkmark}%
}
\usepackage{cuted}
\usepackage{capt-of}

\usepackage[breaklinks=true,letterpaper=true,colorlinks,bookmarks=false]{hyperref}

\cvprfinalcopy 

\def\cvprPaperID{****} 

\ifcvprfinal\fi
\title{VORNet: Spatio-temporally Consistent Video Inpainting for Object Removal}

\author{Ya-Liang Chang \and Zhe Yu Liu \and Winston Hsu \and \\
National Taiwan University, Taipei, Taiwan \\
{\tt\small \{yaliangchang, zhe2325138\}@cmlab.csie.ntu.edu.tw, whsu@ntu.edu.tw}
}
\begin{document}
\maketitle

\begin{strip}\centering
\vspace{-5mm}
\includegraphics[width=\linewidth]{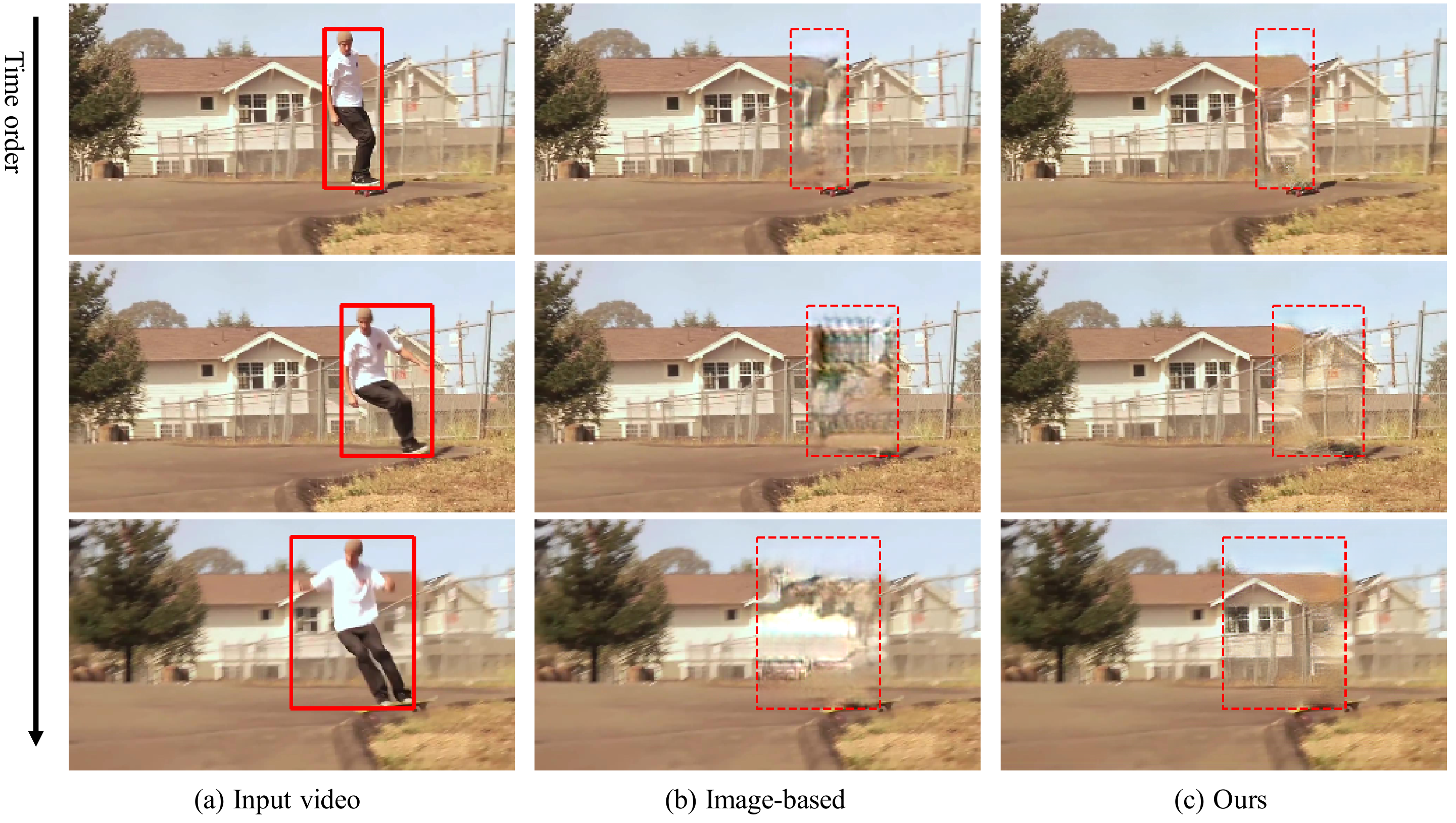}
\captionof{figure}{Video object removal results. (a) The input video and its foreground bounding boxes to remove, marked in red. (b) State-of-the-art image-based inpainting model by Yu \etal \cite{yu2018generative}. (c) Our results. Applying image-based algorithms on the video inpainting task often leads to temporal inconsistency, where the content is different in each frame, \eg the windows are missing in the last frame of (b). Our deep learning based architecture could improve both the spatial and temporal consistency of image-based inpainting models. 
Best viewed with color and zoom-in.
See \url{http://bit.ly/2GkW9Kr} for the video.
\label{fig:teaser}}
\end{strip}

\begin{abstract}

Video object removal is a challenging task in video processing that often requires massive human efforts. Given the mask of the foreground object in each frame, the goal is to complete (inpaint) the object region and generate a video without the target object. While recently deep learning based methods have achieved great success on the image inpainting task, they often lead to inconsistent results between frames when applied to videos. In this work, we propose a novel learning-based Video Object Removal Network (VORNet) to solve the video object removal task in a spatio-temporally consistent manner, by combining the optical flow warping and image-based inpainting model. Experiments are done on our Synthesized Video Object Removal (SVOR) dataset based on the YouTube-VOS video segmentation dataset, and both the objective and subjective evaluation demonstrate that our VORNet generates more spatially and temporally consistent videos compared with existing methods.

\end{abstract}

\section{Introduction}

Removing undesired objects in videos is crucial to many applications, such as movie post-production and video editing. While manually removing objects in a video requires substantial human efforts, automatic video object removal could save a great amount of time. Given the region of the foreground object in each frame, the goal of automatic video object removal is to fill in, or inpaint, the foreground region with background content and generate a video without the target object. Automatic video object removal is a very challenging task since it requires both spatial and temporal consistency; the inpainted region must fit in the background seamlessly in diverse scenes, and it should remain consistent appearance in the following frames where its surroundings may change significantly. Some examples of inconsistent frames include flickering and distortion (see Fig. \ref{fig:teaser}(b) and Fig. \ref{fig:visual_result}(b)(c)).

Video object removal could be viewed as an extension of the image/video inpainting task. Early patch-based inpainting methods \cite{criminisi2003object, criminisi2004region, wu2006object} divide images into small “patches” and recover the masked region by pasting the most similar patch somewhere in the image/video. These methods could generate authentic results but they are usually very time-consuming due to the complexity of neighbor-finding algorithms \cite{le2017motion}. In addition, patch-based methods assume there is a reference for the missing part and often fail to recover non-repetitive and complex region (e.g, they cannot recover a missing face well \cite{li2017generative}).

On the other hand, deep learning based image inpainting models could estimate the missing parts based on the training data and generate novel results with impressive quality \cite{ liu2018image, yan2018shift, yang2017high, yeh2017semantic, yu2018generative}. One naive idea is to solve the video object removal problem by applying these image inpainting models to each frame to recover the foreground region. Nonetheless, when applied to videos, these image-based methods would generate temporally inconsistent results that cause flickering or distorted videos, since they do not consider the temporal relation between frames and treat them independently. 



We propose a novel learning-based architecture for video object removal that could take advantage of existing state-of-the-art image-based inpainting models and generate visually plausible frames in a temporally consistent manner. The core idea is to combine the information from previous frames and generated result in current frame. For previous information, we use the optical flow to capture background motion and recover removed foreground part by warping the previous background accordingly. For the constantly occluded region, existing image-based inpainting models could generate plausible results. Based on these candidates, we design a refinement network to select and refine them to derive a spatially and temporally consistent result. 

Since there is no existing dataset for video object removal, we build a large-scale Synthesized Video Object Removal (SVOR) dataset based on the YouTube-VOS \cite{xu2018youtube} video segmentation dataset. A variety of foreground segmentation and background videos are selected from YouTube-VOS videos and synthesized to 1958 video-with-target and video-without-target pairs. We train our VORNet on the SVOR dataset with reconstruction loss, perceptual loss and two designed GAN losses and evaluate the quality of videos with mean square error, SSIM \cite{wang2003multiscale}, a learned perceptual metric \cite{zhang2018unreasonable} and visual results. We show that the proposed method could improve the perceptual quality and temporal stability. Our VORNet processes frames online, sequentially, does not require post-processing and could deal with videos in various lengths.

Our contributions could be summarized with the following points:
\begin{itemize}
\item We propose a novel Video Object Removal Network (VORNet) to remove undesired objects in videos. To our knowledge, VORNet is the first learning based model for video object removal. It could generate visually plausible and temporally coherent result online, without post-processing.
\item We design a combination of spatial content losses and temporal coherent loss based on GAN structure to train our model, which could improve the spatio-temporal quality of generated videos.
\item We create the first large-scale Synthesized Video Object Removal (SVOR) dataset based on the YouTube-VOS dataset. The SVOR dataset contains a huge variety of motions and scenes that could be used for training and evaluation in further research. The dataset is publicly available here: \url{http://bit.ly/2P3n2oH}.
\end{itemize}

\section{Related Work}

\paragraph{\textbf{Image Inpainting}}
Image inpainting was first introduced in \cite{bertalmio2000image} as a general image processing problem that aims to recover the damaged or missing region of an image. Subsequently, a great amount of research is done for image inpainting \cite{guillemot2014image} with diffusion-based \cite{bertalmio2001navier, bertalmio2000image} and  patch-based \cite{bornard2002missing, criminisi2003object, drori2003fragment, wexler2007space} algorithms. These traditional methods perform well on simple structure but are very limited to complex objects, large missing area and non-repetitive texture where similar reference may not exist.

In recent years, learning-based models demonstrate promising results with the help of deep convolutional neural network (CNNs). These models learn image features in the training data and are thus capable of generating realistic content that may not exist in the unmasked area, such as faces \cite{li2017generative, yu2018generative}, complex objects \cite{miyato2018spectral} and natural scenes \cite{iizuka2017globally, yu2018generative}. Xie \etal \cite{xie2012image} is the first to train convolutional neural networks for image denoising and inpainting on small regions. Pathak \etal \cite{pathak2016context} further extend the work to a larger region by an encoder-decoder structure. Also, to improve blurry effect caused by the $l_2$ loss, Pathak \etal \cite{pathak2016context} introduce the idea of adversarial loss from the generative adversarial network (GAN) \cite{goodfellow2014generative} where a generator that aims to create real images to fool the discriminator and a discriminator that strikes to tell the fidelity of generated images are jointly trained.

More recently, Yu \etal \cite{yu2018generative} add a contextual attention layer to and several improvements on network design to produce higher-quality images. It is trained on the diverse Places2 \cite{zhou2017places} dataset and achieve state-of-the-art result, so we shall take it as our inpainting network and a baseline. 

Yan \etal \cite{yan2018shift}, Yu \etal \cite{yu2018free} and Lui \etal \cite{liu2018image} also manage to solve the problem of inpainting irregular holes. However, since precise segmentation of an object in a video may not be derived easily, we focus only on inpainting bounding box region of the object in this work. We assume the foreground bounding boxes are given as they could be easily derived by object tracking methods or human annotations.

\paragraph{\textbf{Video Inpainting}}
Video inpainting is generally viewed as an extension of the image inpainting task with larger search space and temporally consistent constraints. 
Early works \cite{granados2012not, wexler2007space, newson2014video} are mainly extensions of patch-based methods from image inpainting, where images are split into small patches and the masked region is recovered by pasting the most similar patch somewhere in the image/video. Wexler \etal \cite{wexler2007space} consider the video inpainting task as a global optimization problem that all missing portion could be filled in with patches from the available parts of the video with enforced global spatio-temporal consistency.  Wexler \etal \cite{wexler2007space}  propose an iterative approach to solve the global optimization problem and yield magnificent results in an automatic way. However, due to the large search space and the complexity of the nearest neighbor search algorithm, their method is extremely slow that processing a few seconds of video may take days to compute. 
Also, the assumption that there exists a similar patch that could fill in the missing region may not hold under circumstances like a long-lasting occlusion, a moving camera or masked regions with semantic ambiguity. \cite{ilan2015survey}.

The following works try to solve these issues. Newson \etal \cite{newson2014video} extend the work of Wexler \etal \cite{wexler2007space} by accelerating the algorithm, adding texture features and initialization scheme. Ebdelli \etal \cite{ebdelli2015video} also limit the search space in an aligned group of frames to reduce computational time. Huang \etal \cite{Huang-SigAsia-2016} address the moving camera problem by estimating the optical flow and color in the missing regions jointly. However, the computation time of these methods is still longer than per-frame processing after acceleration. In addition, patch based models still lack modeling distribution of real images, so they fail to recover unseen parts in the video. Our data-driven method could solve both issues by learning the distribution of frames and generate realistic videos by forward inference, without searching.



\begin{figure*} [ht]
\begin{center}
\includegraphics[width=\linewidth]{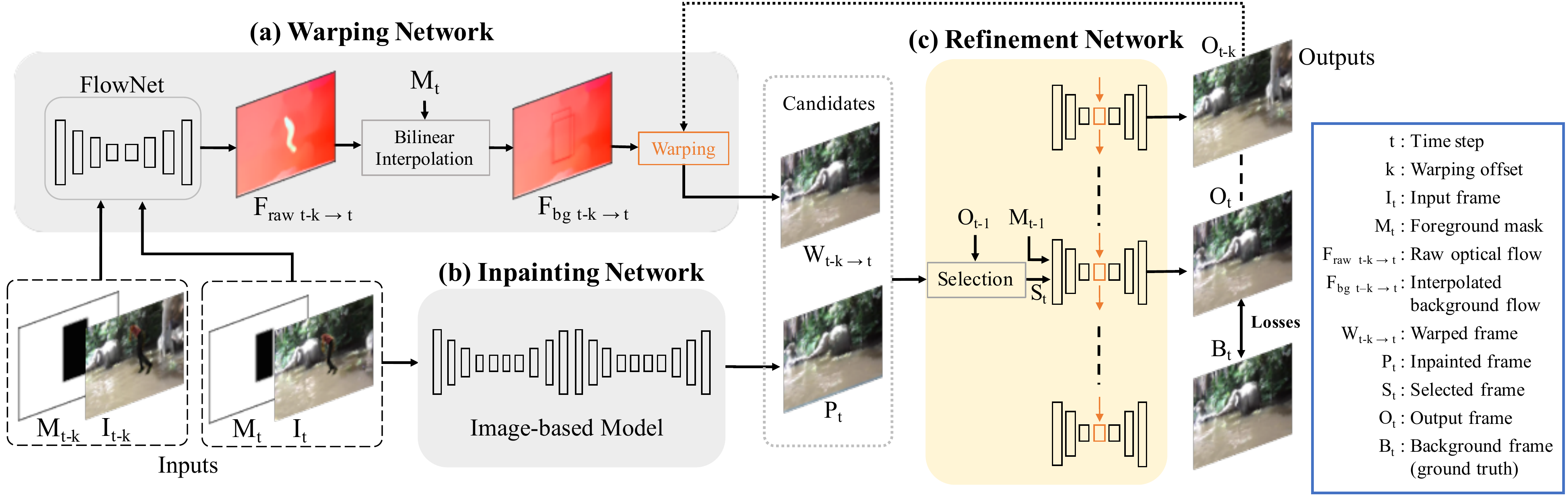}
\end{center}
\caption{Our VORNet architecture and notations. 
(a) The warping network aims to collect information from other frames (see details in Fig. \ref{fig:warping_concept}). (b) The inpainting network intends to estimate the missing parts, by using the generative model to create a possible image according to its surroundings. (c) The refinement network is designed to combine the information from other frames and the estimated frame, by selecting and refining candidates (see details and losses in Fig. \ref{fig:refinement_network}). The model runs in a recurrent way; the output frame is used to create warping candidates of the next frame, and the current state in refinement network would propagate to the next frame. Best viewed with color.}
\label{fig:model}

\end{figure*}

\paragraph{\textbf{Video Temporal Consistency}}
Video temporal consistency aims to solve the flickering problem when applying different kinds of image-based models like photo enhancement \cite{chen2018deep} colorization \cite{zhang2017real}, style transfer \cite{huang2017arbitrary, luan2017deep} and general image-to-image translations \cite{isola2017image, zhu2017unpaired} to videos. Generally, it could be divided into task-independent and task-specific methods. 

Task-independent approaches \cite{bonneel2015blind, Lai-ECCV-2018, yao2017occlusion} aim to use a single model to handle multiple applications with the video temporal consistency problem. Among them, the recent work by Lai \etal \cite{Lai-ECCV-2018} propose an efficient method using a deep network that could generate impressive temporally coherent videos in real time, given various types of temporally inconsistent inputs and their original unprocessed videos as reference. They use the FlowNet2 \cite{ilg2017flownet, flownet2-pytorch} to estimate their temporal loss to train the model. However, for the video object removal task, the method of Lai \etal \cite{Lai-ECCV-2018} does not work because the inpainted region in the unprocessed video is occupied by the foreground object, which not be used as a reference for temporal loss. Instead, our refinement network utilizes the warping network and temporal discriminator to generate temporal consistent results.

Task-specific approaches like \cite{chen2017coherent, levin2004colorization, ye2014intrinsic} develop different strategies according to each domain. Some attempt to design specific temporal filters \cite{aydin2014temporally} or embed optical flow estimation to capture information of motion \cite{chen2017coherent}. Recently, Xie \etal \cite{xie2018tempoGAN} design a temporal discriminator aside from a normal spatial one for the fluid flow super-resolution task. It utilizes motion from low-resolution video to generate temporally consistent high-resolution fluid flow video, but there is no such reference in the video inpainting task. Alternatively, we extend this work to design our temporal discriminator without reference for video inpainting. 

Wang \etal \cite{wang2018videoinp}, concurrent with our work, propose a deep learning architecture to address the inconsistent problem in video inpainting. The method uses a 3D convolutional network to learn the temporal relation and generate coarse temporally consistent images for the masked area, and refine them with a 2D convolutional network. Although results of Wang \etal \cite{wang2018videoinp}  are temporally consistent, their model could not generate clear videos for a diverse dataset as only the $L_1$ loss is used for training. Our VORNet could utilize existing image-based inpainting models and improve the video quality by the combinations of different loss functions .


\section{Video Object Removal}
Our VORNet takes as input the video-with-target frames $\{I_t  \mid  t=1 \dots n\}$  and the target bounding box mask in each frame $\{M_t  \mid  t=1 \dots n\}$ in sequence and generate the output video-without-target frames $\{O_t  \mid  t=1 \dots n\}$. The model is composed of three parts: the warping network, the inpainting network, and the refinement network (see Fig. ~\ref{fig:model}). The core concept is to use the information from other frame (warping network) and generated frame (inpainting network), combine and refine them in a spatio-temporally coherent way (refinement network).

\subsection{The Warping Network}
\begin{figure}[ht]
\begin{center}
\includegraphics[width=\linewidth]{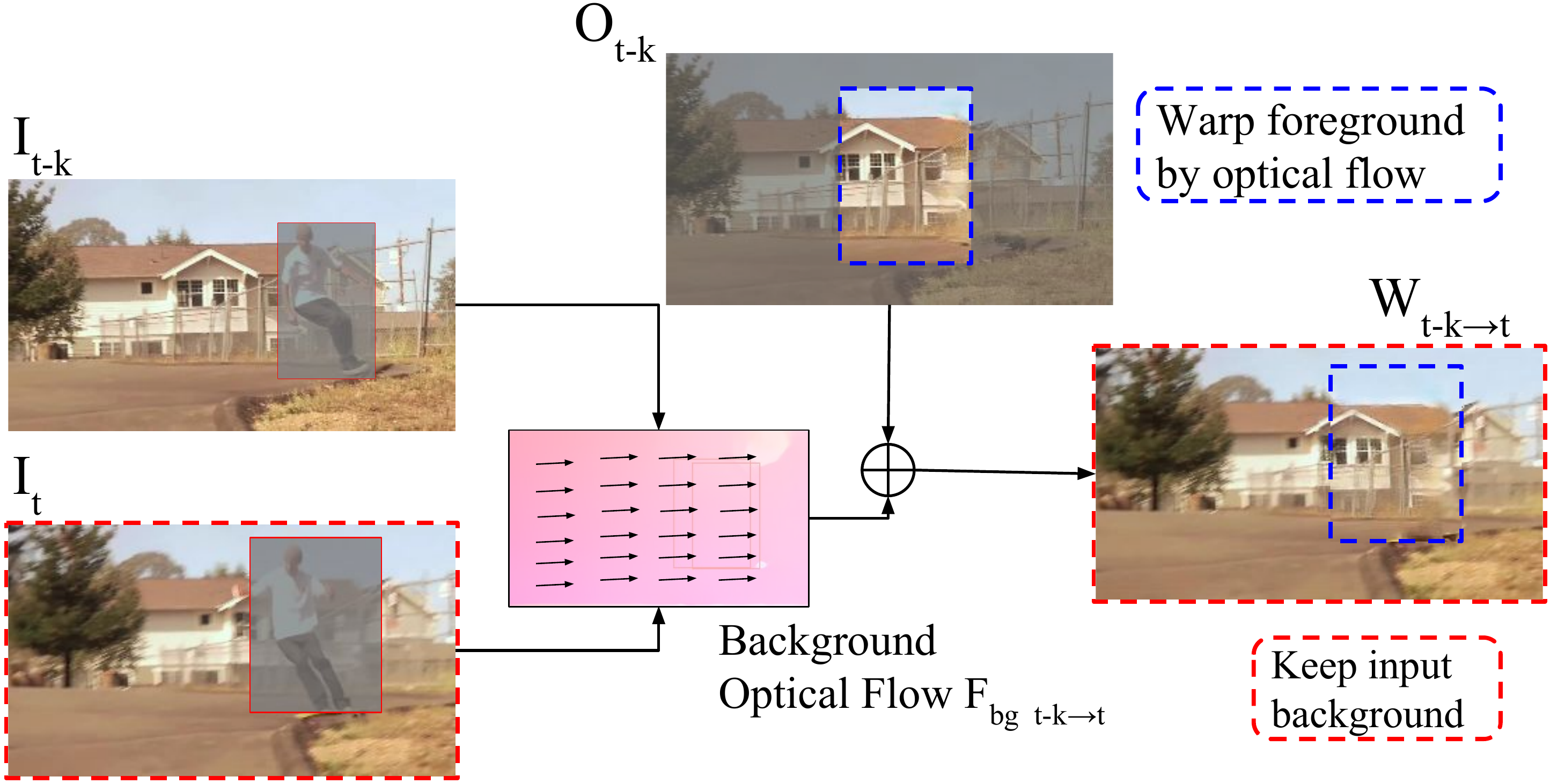}
\end{center}
\caption{Concept of the warping network. In the current input frame $I_t$, the window is hard to be reconstructed for image-based models since it is occluded by the man, but we could easily know that the window should be there seeing the previous frame $I_{t-k}$. Based on this idea, the warping network estimates the motion (optical flow) between the $I_t$ and its $k$th previous frame $I_{t-k}$, and generate the warped frame $W_{t-k \rightarrow t}$ by warping the corresponding foreground area in the previous output $O_{t-k}$ (marked in blue) to $I_t$. Note that we use $O_{t-k}$ to warp the region instead of $I_{t-k}$ because $I_{t-k}$ may include the foreground. Best viewed with color and zoom-in.}
\label{fig:warping_concept}
\end{figure}

The warping network aims to collect information from other frames. For example, if the target object to be removed is static in two consecutive frames and the background is moving rightward, we could know that the foreground region of the second frame should be filled in with the background in its left side in the first frame (see Fig. \ref{fig:warping_concept}). 

To estimate these relative motions between two input frames $I_{t-k}$ and $I_t$, we use FlowNet2 \cite{ilg2017flownet, flownet2-pytorch} pre-trained on the MPI-Sintel Dataset \cite{Butler:ECCV:2012}  to calculate the raw optical flow $F_{{raw}_{t-k \rightarrow t}}$ between them. 

However, for $F_{{raw}_{t-k \rightarrow t}}$, the foreground region is derived from the pasted foreground object, which could not represent the background motion between the last frame and this frame. To address this issue, we remove the foreground region in the raw optical flow and apply simple bilinear interpolation to fill in the removed region and recover the background optical flow $F_{{bg}_{t-k \rightarrow t}}$. We do not adopt the learning based method in this component because the performance is not as expected considering its cost.

Finally, we warp $O_{t-k}$ to the $M_t$ region with inpainted flow $F_{{bg}_{t-k \rightarrow t}}$ using the warping operation as \cite{sun2018pwc} and send it to the refinement network as a candidate. We have candidates with different $k$ so that we could get information from temporally closer and further neighbors.

\subsection{The Inpainting Network}
The inpainting network intends to estimate the missing background. It could be any model that recover the masked part of input videos, including learning based and patch based ones. To estimate occluded regions that patch-based models could not handle, we adopt the generative inpainting network from Yu \etal \cite{yu2018generative} pre-trained on the Places2 dataset \cite{zhou2017places} and fine-tune on our SVOR dataset. It consists of a coarse network that generates a coarse result from the masked input image and a refinement network that turns the coarse result to the final output with contextual attention. Details for the model could be found in the supplementary material.

\begin{figure*}[ht]
\begin{center}
\includegraphics[width=\linewidth]{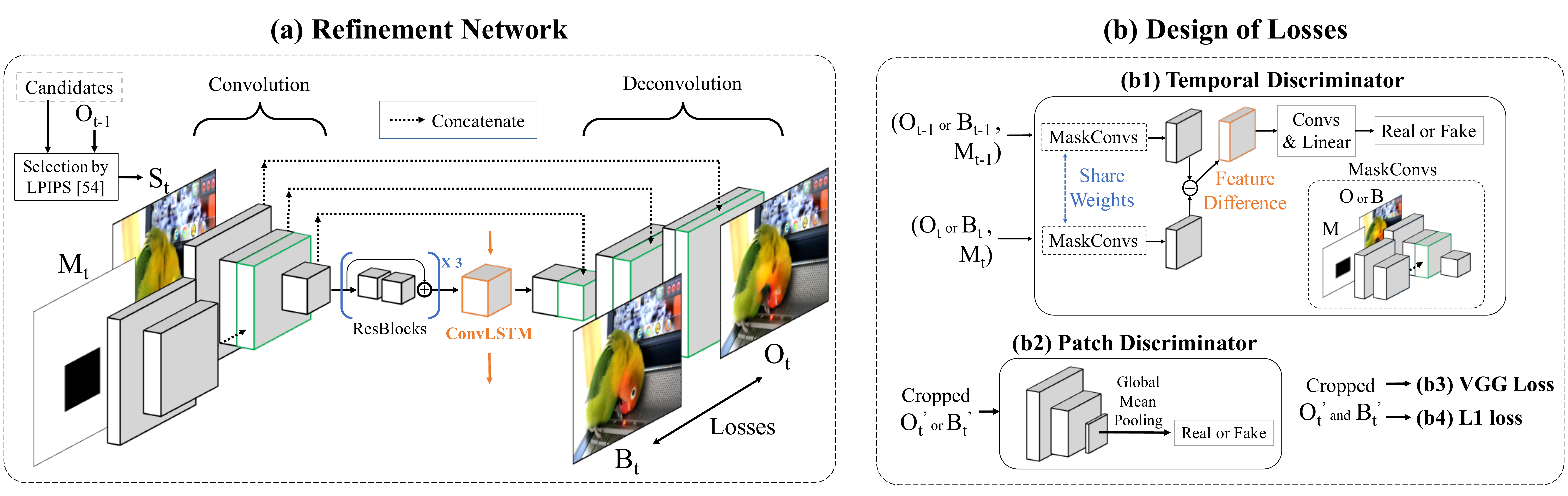}
\end{center}
\caption{Refinement network architecture and losses design. (a) The network select the $S_t$ from candidates (warped and inpainted frames) by the closest LIPIS \cite{zhang2018unreasonable} distance to the last output, and use $S_t$ and the foreground mask $M_t$ to generate the final output $O_t$. The temporal information is propagated by the convolutional LSTM (ConvLSTM) \cite{xingjian2015convolutional} in high-level features. Lastly, losses between the output $O_t$ and the ground truth background $B_t$ will be computed. (b1) The temporal discriminator will estimate the real/fake of output or ground truth frames by the siamese features \cite{chopra2005learning} differences with the last frame. (b2) PatchGAN focus only on masked area. Best viewed with color and zoom-in.}
\label{fig:refinement_network}
\end{figure*}

\subsection{The Refinement Network}
The refinement network is designed to combine the candidates from warping network and inpainting network. Given candidates from warped frames and the inpainted frame, the refinement network will select the top 1 candidate $S_t$ to generate the final output frame $O_t$ with the mask $M_t$. To maintain temporal consistency, the selection is done by choosing the candidate that is closest to the previous result in the feature level (LPIPS \cite{zhang2018unreasonable} distance, see Sec \ref{LPIPS}). Finally, losses are computed using the output $O_t$ and the background frame $B_t$. 

As shown in Fig. \ref{fig:refinement_network}, the refinement network includes three convolutional layers that encode the candidate frames and mask, a convolutional LSTM \cite{xingjian2015convolutional} layer that propagates temporal features, and three transposed convolutional layers to reconstruct the image. Skip connections are added between convolutional layers and corresponding transposed convolutional layers. 

\subsection{Loss Functions}
Our aim is to recover the background region which is masked by the foreground object. This is a challenging task because we need to consider both the spatial and temporal consistency. Accordingly We propose to train our VORNet with spatial content loss from low-level to high-level and temporally coherent loss.

\paragraph{\textbf{Spatially discounted reconstruction loss.}}
 $l_1$ loss
focuses on the lowest level of pixel difference.  We embrace the spatially discounted reconstruction loss in \cite{yu2018generative}
\begin{equation}
L_{l_1} = \mathds{E}_{t,x,y}[ |O_{t_{x,y}} - B_{t_{x,y}}| (\gamma_{t_{x,y}})^d]
\end{equation}
where $x$ and $y$ denote the pixel indexes of a frame, and the loss in each pixel is weighted by $\gamma^d$ according its distance $d$ to the nearest boundary. It is more suitable for image inpainting task compared to naive $l_1$ loss since pixels closer to the boundary should match the background, while the middle part could have more diversity.

\paragraph{\textbf{VGG perceptual loss.}}
One problem about $l_1$ loss in a generative task is that it usually produces blurry results, because it is hard for the model to minimize the $l_1$ loss when generating a sharp and vivid image. 

Therefore, we adopt a VGG-net pre-trained on a classification task \cite{johnson2016perceptual} to compute the perceptual distance between generated and ground truth images as one of our spatial loss
\begin{equation}
L_{perc}^{\phi, j} = \mathds{E}_t[\| \phi_j({O'}_t) -\phi_j({B'}_t) \|^2_2]
\end{equation}
where $\phi$ and $j$ denote the VGG network and its layer index respectively. The ${O'}_t$ and ${B'}_t$ are the output and background image cropped to the masked area. The perceptual loss emphasizes on the higher level of difference like style or textures instead of pixels.

\paragraph{\textbf{PatchGAN loss.}} To motivate our model to generate realistic images, we use the PatchGAN discriminator \cite{isola2017image} as our spatial discriminator $D_s$, while the refinement model could be viewed as a generator $G$. The Patch GAN loss is defined as
\begin{equation}
\begin{aligned}
L_{GAN_s}(G, D_s) &= \mathds{E}_t[log(D_s({B'}_t))] \\                     &+ \mathds{E}_t[log(1-D_s(G(S'_t)))]
\end{aligned}
\end{equation}
where ${S'}_t$ and ${B'}_t$ denote the selected and background image cropped to the masked area.
While the $l_1$ loss focus on low-frequency structure, PatchGAN discriminator penalizes local patches only for the high-frequency structure \cite{isola2017image}.

\paragraph{\textbf{Temporal GAN loss.}}
The above losses are for the image quality only, while we need a temporal constraint to generate content coherent videos. 
To solve this problem, we design a temporal discriminator to train our model. 
A similar idea could be seen in a recent fluid flow super-resolution work \cite{xie2018tempoGAN}, which propose the TempoGAN to generate temporally coherent high-resolution fluid flow video utilizing flow motion in low-resolution one. 
While for video inpainting there is no low-resolution reference, our temporal discriminator estimates the consistency score by the differences of consecutive frames in the feature level. It takes the features from output frames and the foreground masks as inputs, calculates siamese features \cite{chopra2005learning} differences,
 further extract features and estimate the final consistent score (see Fig. \ref{fig:refinement_network}b1). With the proposed temporal discriminator $D_t$, the temporal GAN loss is defined as:
\begin{equation}
\begin{aligned}
L_{GAN_t}(G, D_t) &= \mathds{E}_t[log(D_t(B_t))] \\              &+ \mathds{E}_t[log(1-D_t(G(S_t)))]
\end{aligned}
\end{equation}

\paragraph{\textbf{Overall loss.}}
The overall loss function to train our VORNet is defined as:
\begin{equation}
\begin{aligned}
L &= \lambda_{l_1} \times L_{l_1}    +
\lambda_{perc} \times L_{perc}^{\phi, j}   \\      &+
\lambda_{G_s} \times L_{G_s}     +
\lambda_{G_t} \times L_{G_t}
\end{aligned}
\end{equation}
where $\lambda_{l_1}$, $\lambda_{perc}$, $\lambda_{G_s}$ and $\lambda_{G_t}$ are the weights for reconstruction loss, perceptual loss, spatial GAN loss and temporal GAN loss, respectively.

\begin{figure*}
\begin{center}
\includegraphics[width=\linewidth]{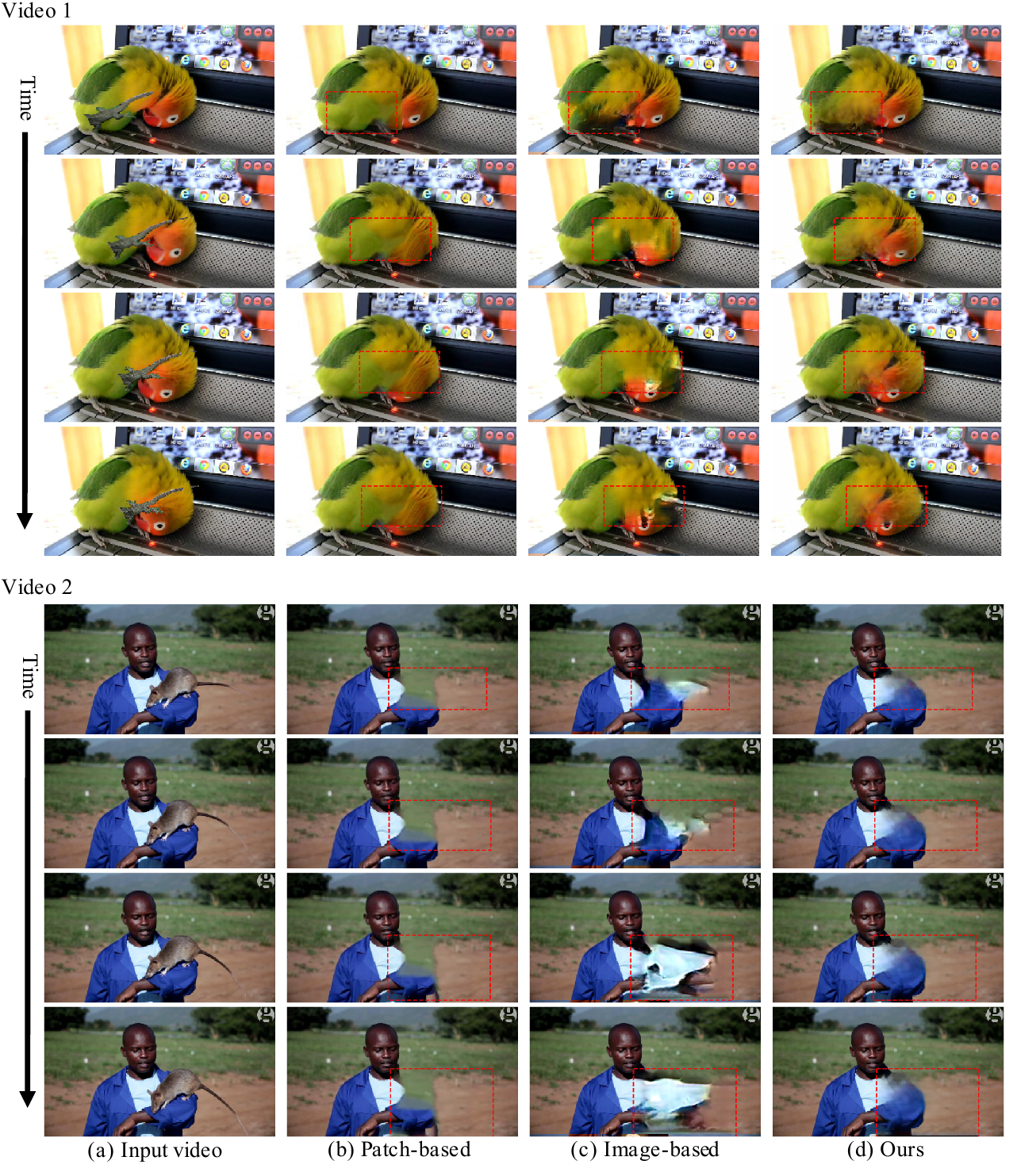}
\end{center}
\caption{\textbf{Visual results} compared with (b) state-of-the-art patch-based video inpainting by Huang \etal \cite{Huang-SigAsia-2016} and (c) state-of-the-art image-based inpainting by Yu \etal \cite{yu2018generative}. \textbf{Video 1}: the first frame is spatially consistent for all methods. However, for (b) and (c), the bird lose its eye, while ours keeps it intact by the warping network. \textbf{Video 2}: the man's shoulder is filled with grass for the patch-based method (b) as it could not tell the surroundings. For image-based method (c), we can see that the second and third frame is very different, while our results remain temporally consistent.
Best viewed in color and zoom-in.}
\label{fig:visual_result}

\end{figure*}

\section{Experimental Results}
\subsection{Dataset}
We build our Synthesized Videos for Object Removal (SVOR) dataset based on the YouTube-VOS dataset \cite{xu2018youtube}, which is a large-scale dataset for video segmentation including a huge variety of moving objects, camera view and motion types. The YouTube-VOS dataset consists of 4,453 videos and 7,822 unique objects including humans with diverse activities, animals, vehicles, accessories and some common objects. Each video is about 3 to 5 seconds, and up to five human annotated object segmentation masks are given every five frames in a 30 FPS frame rate.

Since it is not likely to get the real ground truth for the video object removal task 
, we utilize segmentation in the YouTube-VOS training set to synthesize training video-with-target/video-without-target pairs. After manually filter out videos where the annotated object is only partially in the screen, occupied more than one-half of the screen or smaller than $30 \times 30$ pixels, 1,958 videos are used to synthesize our input videos. 

We split these videos into 1,858 training and 100 testing videos, and create 1,858 and 100 pairs among them (there could be significantly more pairs if existing duplicated foreground/background videos). Each synthesized video pair is composed of one foreground video and one background video from the YouTube-VOS dataset. We take the first object segmentation mask in the foreground video as the target objects and paste it to the background video. Consequently, it becomes the input synthesized video-with-target, and the background video is viewed as the ground truth video-without-target.

In the SVOR dataset, the foreground object may be static or moving, and its size could vary in a single video. Also, the background may be shaky, following an object or changing the brightness.
Some background objects could also be originally in the foreground region or moving toward the region, so the SVOR dataset is very diverse and challenging.


\subsection{Implementation Details}
Our model is implemented with Pytorch 0.4.1 and trained on our SVOR dataset in $320 \times 180$, at most 15 frames for each video. Warping temporal distance $k$s are set to be \{1, 3, 5\}. The $relu3_3$ layer is used for the VGG loss. $\gamma$ for $L_{l_1}$ is set to be 0.99. The patch size for PatchGAN is $15 \times 15$. Loss weights $\lambda_{l_1}$, $\lambda_{perc}$ and $\lambda_{G_s}$ are set to be 1. $\lambda_{G_t}$ is 0.01. Other details could be found in the supplementary material.

\subsection{Quantitative and Qualitative Comparisons}

We compare the proposed method with the well-known patch-based video inpainting methods \cite{newson2014video, Huang-SigAsia-2016} with patch size $3 \times 3 \times 3$, the state-of-the-art image-based inpainting model \cite{yu2018generative} pre-trained on the Places2 dataset and fine-tuned on our SVOR dataset and the two-stage learning based video inpainting model \cite{wang2018videoinp}. In general, our VORNet performs better than the four benchmarks quantitatively and qualitatively.

We report the evaluation in terms of mean square error (MSE) and structural similarity (SSIM) \cite{wang2003multiscale}, which are commonly used in inpainting tasks \cite{liu2018image, yan2018shift, yu2018generative}. However, these traditional evaluation metrics may not represent the perceptual distance well (i.e., they prefer blurry images than partially shifted, distorted images).  As a result,  we also use the recently proposed Learned Perceptual Image Patch Similarity (LPIPS) \cite{zhang2018unreasonable} \label{LPIPS} to estimate perceptual distance. LPIPS calibrates features of ImageNet classification networks and corresponds more to human perception. We take the model calibrated on the AlextNet \cite{krizhevsky2012imagenet} as suggested \cite{zhang2018unreasonable}. The quantitative result of 100 synthesized video in the testing set could be seen in Table. ~\ref{tab:quantitative_benchmark}. We could see that our model outperforms the four benchmarks.

Since the quantitative result of frames may not represent the temporal consistency, we also evaluate qualitatively on 100 testing videos. The visual comparison could be seen in Fig. ~\ref{fig:visual_result}. Our results remain spatio-temporally stable as surroundings change, while results of other methods become distorted or inconsistent. More visual comparisons with all baselines \cite{newson2014video, Huang-SigAsia-2016, yu2018generative, wang2018videoinp} could be found in \url{https://bit.ly/2I7WbID} (synthesized), \url{https://bit.ly/2GdnbUX} (real) and the supplementary material.

\subsection{Ablation Study}

To evaluate the contribution of each component in the proposed model, we conduct ablation study on main components including the warping network, VGG loss, spatial discriminator and temporal discriminator. The result is shown in Table. ~\ref{tab:ablation_study}. We could see that the warping network play an important role in our VORNet, while each loss has some effects on the result. Specifically, if VGG loss is removed, the model would generate sharp images disregarding the surrounding content; if the spatial GAN loss is removed, there would be some unnatural repetitive patterns that could reduce MSE; if the temporal GAN loss is removed, the result would be slightly temporally inconsistent. Corresponding visual comparisons could be found in the supplementary material.

\subsection{Execution Time}

The execution time is evaluated on a machine with a Intel Xeon E5-2650 v3 CPU (128G RAM) and two Nvidia Tesla K80 GPUs. The speed of VORNet is 2.5 frame per second (FPS), slower than the Yu \etal \cite{yu2018generative} (11 FPS) due to FlowNet2 \cite{ilg2017flownet, flownet2-pytorch} full-model optical flow estimation (ours is 7 FPS with FlowNet2-S \cite{ilg2017flownet, flownet2-pytorch}), while faster than the video inpainting method \cite{newson2014video} (0.15 FPS) with patch size $3 \times 3 \times 3$ since it runs on the CPU. Note that our VORNet does not require post-processing and can run online, without peeking the future frames.

\subsection{Limitations and Discussion}
Our model relies on the optical flow to get information from the previous frames, which results in extra execution time and parameters. In addition, the state-of-the-art networks for optical flow inference still could not capture object motions in detail and there is unavoidable occlusion problem, which make the warped frames blurry. 
A possible solution is to design a temporal attention and warping network that could replace the optical flow warping. The model could be trained in an end-to-end way and the performance may be improved. Still, the proposed method is the first learning-based architecture for the video object removal task and produces state-of-the art results. 

\begin{table}[]
\begin{tabular}{llll}
\hline
Method  & MSE   $\downarrow$   & SSIM    $\uparrow$   & LPIPS  $\downarrow$     \\ \hline
Huang \etal \cite{Huang-SigAsia-2016} & 0.01665       & 0.6967     & 0.2385          \\
Newson \etal \cite{newson2014video} & 0.02152       & 0.6577     & 0.2409          \\
Yu \etal \cite{yu2018generative}    & 0.02009      & 0.6896     & 0.2249          \\
Wang \etal \cite{wang2018videoinp}    & 0.01566      & 0.6749     & 0.3915          \\
VORNet (Ours)  & \textbf{0.01560} & \textbf{0.7260} & \textbf{0.1889} \\ \hline
\end{tabular}

\caption{Quantitative results of the proposed network, state-of-the-art patch-based video inpainting \cite{Huang-SigAsia-2016, newson2014video}, image inpainting \cite{yu2018generative} and learning-based video inpainting \cite{wang2018videoinp}  methods. We could use original background videos as ground truth to calculates these metrics since we evaluate on our synthesized dataset. 
}
\label{tab:quantitative_benchmark}

\end{table}

\begin{table}[]
\begin{tabular}{cccccc}
\hline
\begin{tabular}[c]{@{}c@{}}Warping \\ network\end{tabular} & 
\begin{tabular}[c]{@{}c@{}}VGG \\ loss  \end{tabular} & 
\begin{tabular}[c]{@{}c@{}}Spat. \\ Disc.\end{tabular} & \begin{tabular}[c]{@{}c@{}}Temp. \\ Disc.\end{tabular} & 
MSE  $\downarrow$ & LPIPS$\downarrow$ \\ \hline
 &  $\boldcheckmark$ &   $\boldcheckmark$    &  $\boldcheckmark$  & 0.01807  &  0.2576   \\
$\boldcheckmark$  &   &   $\boldcheckmark$  &  $\boldcheckmark$  & 0.01846   &    0.2460 \\
$\boldcheckmark$  &  $\boldcheckmark$ &         &  $\boldcheckmark$ & \textbf{0.01314}   &   0.2291    \\
$\boldcheckmark$  & $\boldcheckmark$  & $\boldcheckmark$    &   & 0.01669     & 0.2039 \\
$\boldcheckmark$ & $\boldcheckmark$  & $\boldcheckmark$  & $\boldcheckmark$  & 0.01560  & \textbf{ 0.1889}\\
 \hline
\end{tabular}
\caption{Ablation study of the components including warping network, VGG loss, spatial discriminator and temporal discriminator. MSE and LPIPS \cite{zhang2018unreasonable} distance  with the ground truth is calculated for the 100 testing pairs.}
\label{tab:ablation_study}

\end{table}

\section{Conclusion}
In this work, we propose a novel Video Object Removal Network (VORNet) for the video object removal task, utilizing existing image-based inpainting model and enhance the spatial and temporal consistency. To our knowledge, our VORNet is the first to introduce learning-based method to the video object removal task. We design spatial and temporal GAN losses and train the proposed model on our Synthesized Video Object Removal Dataset (SVOR) based on the YouTube-VOS video segmentation dataset. Our model is learning based, runs online, faster than patch-based video inpainting method and does not require post-processing. Evaluation on perceptual distance, visual result and user studies show that our model achieves state-of-the-art results compared to existing methods.

\section*{Acknowledgement}
This work was supported in part by the Ministry of Science and Technology, Taiwan, under Grant MOST 108-2634-F-002-004. We also benefit from the NVIDIA grants and the DGX-1 AI Supercomputer. We are grateful to the National Center for High-performance Computing.

{\small
\bibliographystyle{ieee_fullname}
\bibliography{egbib}
}

\end{document}